\documentclass{hld2026}
\usepackage{multicol}  
\usepackage{float}

\usepackage{graphicx}
\usepackage{amsmath}
\usepackage{amssymb}
\usepackage{bm}
\usepackage{microtype}
\usepackage{booktabs}
\usepackage[labelfont=bf, labelsep=colon]{caption}
\title{Noise-Driven Escape from Metastable Phases explains  Grokking in Deep Neural Networks}

\author{\Name{Ibrahim Talha Ersoy} 
\Email{talha.ersoy@uni-potsdam.de} \\
        \Name{Karoline Wiesner} \Email{karoline.wiesner@uni-potsdam.de} \\
        \addr Complexity Science Group, Institute of Physics and Astronomy,\\
        University of Potsdam, Potsdam, Germany}

\begin{document}

\maketitle

\begin{abstract}
Deep neural networks (DNNs) exhibit first-order phase transitions under variations of the L2 regularisation strength, with each transition marking the onset of a new learnable feature. Below a critical regularisation strength, all features are in principle learnable, but coexisting metastable states, separated by energy barriers, can trap the network and impede convergence. A strength of DNNs is their ability to generalise. But many open questions remain, among them the origin of so-called grokking: the abrupt, delayed onset of generalisation after prolonged apparent overfitting. We show for linear DNNs that grokking is consistent with hysteresis in first-order L2 phase transitions: using L2 regularisation to engineer deliberate trapping, we demonstrate that a model in a low-accuracy metastable state escapes only when SGD noise drives it across an energy barrier, with escape times following Arrhenius scaling. We reproduce grokking-like delayed convergence across two orders of magnitude in escape time by deliberately trapping models in metastable phases. Using sparse sub-sampling we also reproduce the canonical grokking curve where test error eventually approaches the final training error. Our work suggests that the number of metastable states equals the number of learnable features -- one per singular value of the data covariance -- the potential for hysteresis grows naturally with task complexity. We provide evidence that the same mechanism likely operates in general nonlinear DNNs. Our results provide routes toward more efficient learning schemes.

\end{abstract}

\section{Introduction}

%-----------------------------------------------------------------------

L2 regularisation is a cornerstone of modern machine learning, employed to combat overfitting from classical ridge regression \cite{hoerl1970ridge} to large-scale deep learning \cite{Goodfellow-et-al-2017}. Beyond its practical role, L2 regularisation gives rise to rich phenomenology recently understood through statistical physics. Ziyin and Ueda \cite{ziyin2023zeroth} showed that varying regularisation strength drives genuine phase transitions in DNNs, identifying first-order transitions in DNNs at the onset of learning (the transition from under- to over-parameterisation). Subsequent work extended this picture beyond the onset: Ersoy and Wiesner linked these transitions to curvature drops in the loss landscape \cite{ersoy2025curvature}, Ersoy, Licha, and Wiesner connected them to learnable feature hierarchies \cite{ersoy2025hierarchical}, and Ladewig, Ersoy, and Wiesner showed that in linear networks each non-zero singular value of the data covariance produces its own rank transition \cite{ladewig2025linear}, meaning transitions recur throughout training, once per learnable feature. \emph{Grokking}, the sudden transition from near-zero to near-perfect generalisation long after the training loss has saturated, has attracted much attention since Power et al. demonstrated it in transformers trained on modular arithmetic~\cite{power2022grokking}. The origin of this phenomenon is not yet fully understood. Liu et al.~\cite{liu2022omnigrok} implicated regularisation as the central driver of grokking. Nanda et al.~\cite{nanda2023progress} identified interpretable representations at the transition, and Tian~\cite{tian2025provable} derived scaling laws for feature emergence. Rubin et al.~\cite{rubin2024grokking} proposed a first-order phase transition analogy involving an entropy barrier. This was then challenged by Zhang et al.~\cite{zhang2026grokkingcomputationalglassrelaxation}, who found no entropy barrier and proposed glassy relaxation instead. Solvable linear models have likewise been shown to grok~\cite{levi2024grokking}, and a distribution shift between training and test data has been identified as a driver of delayed generalisation~\cite{lyu2024dichotomy}. We offer an alternative account. As the central mechanism we suggest \emph{hysteresis} in analogy to the statistical physics of phase transitions: a model initialised in a low-accuracy phase remains there until SGD noise drives it across an energy (loss) barrier into the globally optimal phase. To show that hysteresis is consistent with the grokking phenomenology, we build on the finding that varying the L2 regularisation strength leads to first-order phase transitions. We ask: Can activated escape from the resulting metastable states, rather than the transitions themselves, account for the hallmark features of grokking? To answer this, we use L2 regularisation as a control tool to engineer metastable trapping. We successfully reproduce grokking behaviour, and, furthermore, show that this activated process is governed by Arrhenius-type kinetics~\cite{kramers1940,hanggi1990} with an effective temperature~\cite{mandt2017sgd} $T_{\mathrm{eff}}\propto\eta_{\mathrm{lr}}/B$. Here $\eta_{\mathrm{lr}}$ is the learning rate and $B$ the batch size, making the escape time exponentially sensitive to hyperparameter choices. Our results, established in Sections~\ref{sec:methods}--\ref{sec:results}, support three claims: 
\textbf{(1)}~first-order L2 phase transitions produce $d$ coexisting metastable states (one per learnable feature) whose energy barriers trap models in low-accuracy phases.
\textbf{(2)}~Escape is analogous to a thermally activated process obeying Arrhenius kinetics
with $T_{\mathrm{eff}}\propto\eta_{\mathrm{lr}}/B$; we confirm this numerically  with $R^2=0.991$.
\textbf{(3)}~Deliberate trapping reproduces hallmarks of grokking, i.e. long delay, abruptness, sensitivity to initialisation, and, under sparse sampling, the train/test dissociation, across two orders of magnitude in escape time.

\section{Methods}
\label{sec:methods}

%-----------------------------------------------------------------------
\subsection{L2 Phase Transitions}
%-----------------------------------------------------------------------
We use deep linear networks as our minimal model because their loss landscape is exactly solvable, allowing us to locate all $d$ metastable minima and energy barriers analytically. All qualitative results, saddle-node bifurcations, coexisting phases, Arrhenius escape, survive in nonlinear networks (Appendix~\ref{app:nonlinear}), but the linear case keeps the analysis tractable. In this section we review the key prior results on L2 phase transitions before introducing our escape-time framework. Ladewig et al. \cite{ladewig2025linear} described the precise mechanism for linear DNNs, linking the transitions to the singular values of the data covariance: with input $x$ and output $y$, covariances $\Sigma_{xx}$, $\Sigma_{yy}$, and cross-covariance $\Sigma_{yx}$, the singular values $\eta_i$ of $\Sigma_{yx}$ directly characterise the learnable features in the aligned case ($\Sigma_{xx}=\mathbf{I}$). After the network reaches the 0-balanced subspace, the set of weight configurations where all layers contribute equally to the end-to-end map, the L2-regularised loss decouples into independent terms for each singular value $\lambda_i$ of the end-to-end weight matrix:
\begin{equation}
  \mathcal{L}(\{\lambda_i\},\beta)
    = \frac{1}{2}\sum_{i=1}^{d}(\lambda_i - \eta_i)^{2}
    + \frac{\beta}{2}\sum_{i=1}^{d}\lambda_i^{2/L},
  \label{eq:loss}
\end{equation}
where $L$ is network depth, $d$ is the number of non-zero modes, and $\beta>0$ is the regularisation strength. For depth $L\ge3$, the stationarity condition $\partial\mathcal{L}/\partial\lambda_i=0$ for a single mode reads:
\begin{equation}
  \lambda - \eta + \beta\,\lambda^{2/L-1} = 0.
  \label{eq:stationary}
\end{equation}
As $\beta$ increases through $\beta_c$, the two non-trivial solutions (a stable minimum and an unstable saddle) merge and annihilate in a saddle-node bifurcation (see Appendix~\ref{app:critical_beta}), leaving only $\lambda=0$. Below $\beta_c$, zero and non-zero rank solutions coexist, separated by a finite energy barrier (see Appendix~\ref{app:bifurcation} for the bifurcation diagram). The critical
regularisation strength is:
\begin{equation}
  \beta_c^{(i)} = \frac{1}{1-k}\left(\eta_i\frac{1-k}{2-k}\right)^{2-k},
  \qquad k=\frac{2}{L}.
  \label{eq:beta_c}
\end{equation}
For $\beta>\beta_c^{(i)}$ the metastable minimum vanishes. The equal-loss point $\beta^*_i$ (where the lower and higher rank solutions have equal regularised loss) lies strictly below $\beta_c^{(i)}$ for $L\ge3$, producing hysteresis: a model remains trapped in the lower rank phase even when the higher rank phase is energetically preferred. The ordering $\eta_1>\cdots>\eta_d>0$ gives $d$ distinct bifurcations, one metastable state per learnable feature.

%-----------------------------------------------------------------------
\subsection{SGD as Langevin Dynamics and Arrhenius Escape}
%-----------------------------------------------------------------------
To model escape from metastable states, we note that SGD mini-batch noise injects stochastic fluctuations into the gradient with covariance scaling as $\eta_{\mathrm{lr}}/B$, where $\eta_{\mathrm{lr}}$ is the learning rate and $B$ the batch size. In the overdamped limit, the dynamics maps onto Langevin dynamics with an effective temperature~\cite{mandt2017sgd}:
\begin{equation}
  T_{\mathrm{eff}} \propto \frac{\eta_{\mathrm{lr}}}{B}.
  \label{eq:Teff}
\end{equation}
The mean escape time from a metastable state then follows the Kramers--Arrhenius
law~\cite{kramers1940,hanggi1990}:
\begin{equation}
  \ln\tau = \ln\tau_0 + \frac{\Delta E_{\mathrm{eff}}}{T_{\mathrm{eff}}},
  \label{eq:arrhenius}
\end{equation}
where $\Delta E_{\mathrm{eff}}$ is an effective barrier height absorbing high-dimensional curvature and entropic corrections (see Appendix~\ref{app:effective_barrier}). This yields the falsifiable prediction that $\ln\tau$ is linear in $B/\eta_{\mathrm{lr}}$.

\section{Results}
\label{sec:results}

%-----------------------------------------------------------------------

\subsection{Hysteresis and Trapping Reproduce Grokking}

\begin{figure}[H]
 \centering
 \begin{minipage}{0.48\textwidth}
   \centering
   \includegraphics[width=\linewidth]{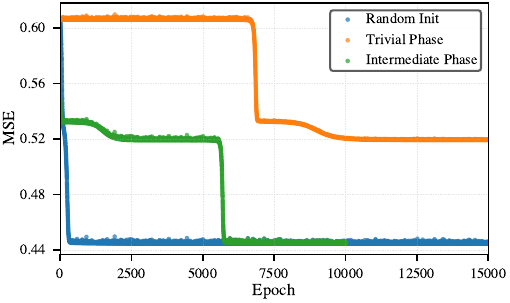}
   \par\vspace{0pt}\small\textbf{(a) Hysteresis as Initialisation Dependence}
 \end{minipage}
 \hfill
 \begin{minipage}{0.48\textwidth}
   \centering
   \includegraphics[width=\linewidth]{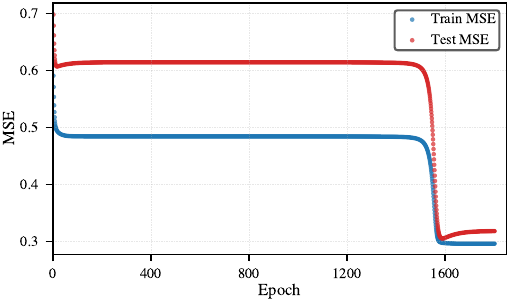}
   \par\vspace{0pt}\small\textbf{(b) Train-Test Discrepancy in Hysteresis}
 \end{minipage}
 \caption{Delayed convergence in deep linear networks. 
   \textbf{(a)} Random (blue), rank‑1 trap (green), trivial‑phase trap (orange) at $\beta=0.03$, $\eta_{\mathrm{lr}}=0.08$, $B=64$. The convergence is strongly delayed when the model is initialised in the local minima of the lower accuracy phases.
   \textbf{(b)} Canonical grokking via sparse sub-sampling ($25$ training samples, $\beta=0.0025$). Initialised in the rank-1 phase, train MSE (blue) drops quickly but only to a plateau while test MSE (red) plateaus higher; at ${\approx}1500$ epochs the model escapes the rank-1 phase and test MSE falls sharply, approaching train MSE to within a small residual gap set by the irreducible noise of the task.}
 \label{fig:mse}
\end{figure}

The coexistence of metastable states implies that the training outcome depends sensitively on initialisation. To demonstrate this we operate at $\beta=0.32$, which lies in the range $\beta<\beta_c^{(1)}$, so the global minimum is rank-2 but a metastable rank-1 minimum also exists. All three experiments use the learning rate $\eta_{\mathrm{lr}}=0.08$ and batch size $B=64$. We consider three initialisation protocols (Fig.~\ref{fig:mse}): \noindent\textbf{(i) Random initialisation.} \emph{Setup}: standard random weight initialisation at $\beta=0.32$. \emph{Observation}: the model converges rapidly to the rank-2 global minimum (Fig.~\ref{fig:mse}, blue; $\tau\approx 10$ epochs).
\emph{Conclusion}: no trapping occurs when the model starts outside any metastable phase. \noindent\textbf{(ii) Rank-1 trap.} \emph{Setup}: model initialised from a checkpoint
trained at $\beta>\beta_c^{(1)}$, placing it in the metastable rank-1 phase. \emph{Observation}: the model remains in the rank-1 state for $\tau\approx 5500$ epochs before abruptly escaping to rank-2 (Fig.~\ref{fig:mse}, green). \emph{Conclusion}: trapping in a metastable phase produces grokking-like delayed convergence.
\noindent\textbf{(iii) Trivial-phase trap.} \emph{Setup}: model initialised from a
checkpoint trained at $\beta>\beta_c^{(2)}$, placing it in the rank-0 phase.
\emph{Observation}: the model escapes to rank-1 after $\tau>7000$ epochs, then remains
trapped in rank-1 for a further extended period, with total delay $\tau>10{,}000$ epochs
(Fig.~\ref{fig:mse}, orange). \emph{Conclusion}: sequential trapping across multiple
metastable phases reproduces staged grokking delays spanning two orders of magnitude.
\noindent\textbf{(iv) Canonical grokking via sparse sub-sampling.}
\emph{Setup}: train and test are drawn from the \emph{same} distribution (weak correlation $0.8$, strong correlation $0.9$), but the training set is very sparse, only $25$ samples ($0.5\%$ of a $5\,000$-sample pool), so the weak feature is poorly determined from the training data while remaining at full strength on test. The model is initialised in the rank-1 trapped state and trained at $\beta=0.0025$, a small regularisation chosen so that the plateau is long-lived but eventually resolves.
\emph{Observation}: After a brief transient in which the strong mode equilibrates, both errors plateau, with train below test (Fig.~\ref{fig:mse}(b)); the model holds in the rank-1 phase for ${\approx}1500$ epochs, then escapes to rank-2, and test error falls sharply to approach train error to within a small residual gap set by the irreducible noise of the stochastic task. \emph{Conclusion}: the same trapping mechanism reproduces the canonical grokking curve when the weak feature is sufficiently underrepresented in the training sample.

\subsection{Energy Barrier Between Metastable States}

The trapping mechanism requires a finite energy barrier between the differing rank phases. Fig.~\ref{fig:barrier_arrhenius}(a) shows the loss landscape section along the minimal-loss path at $\beta=0.32$, parameterised by the singular value $\lambda$. A clear barrier separates the local minimum at $\lambda=0$ from the global minimum. We evaluate Eq.~\eqref{eq:loss} numerically at the saddle point $\lambda^{\mathrm{sad}}\approx0.41$ (obtained from Eq.~\eqref{eq:stationary} at $\beta=0.32$, $\eta=0.8$, $L=3$). This gives $\Delta E_{\mathrm{min}} \approx 0.003$, the barrier along the lowest-loss path out of the metastable phase (Fig.~\ref{fig:barrier_arrhenius}(a)). As we show next, the effective barrier governing escape times is far larger.

\subsection{Arrhenius Scaling Confirms Thermally Activated Escape}
\label{subsec:arrhenius}

\begin{figure}[H]
 \centering
 \begin{minipage}{0.48\textwidth}
   \centering
   \includegraphics[width=\linewidth]{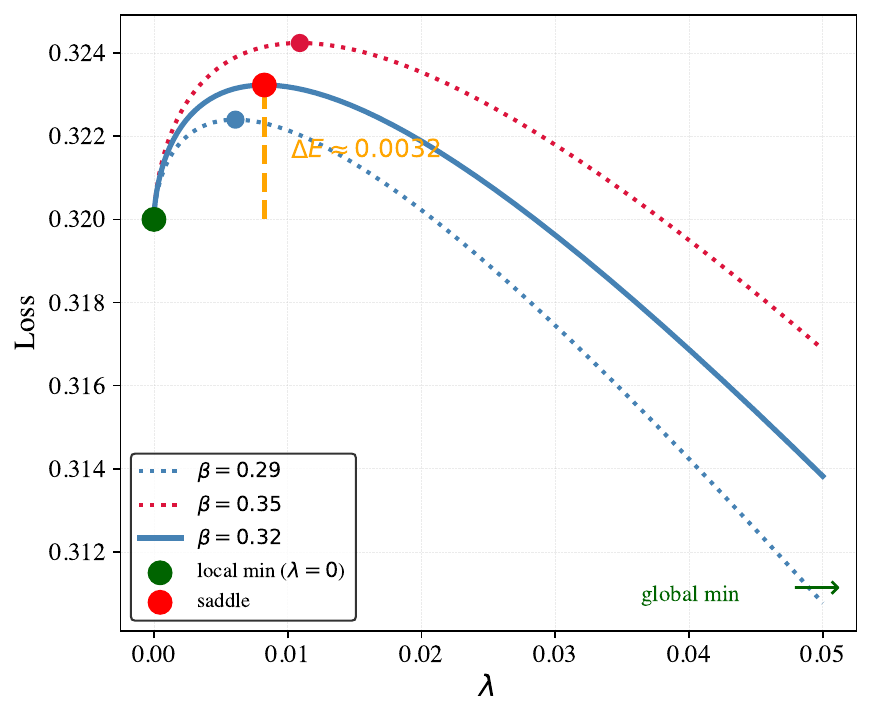}
   \par\vspace{0pt}\small\textbf{(a) Loss Landscape Barrier}
 \end{minipage}
 \hfill
 \begin{minipage}{0.48\textwidth}
   \centering
   \includegraphics[width=\linewidth]{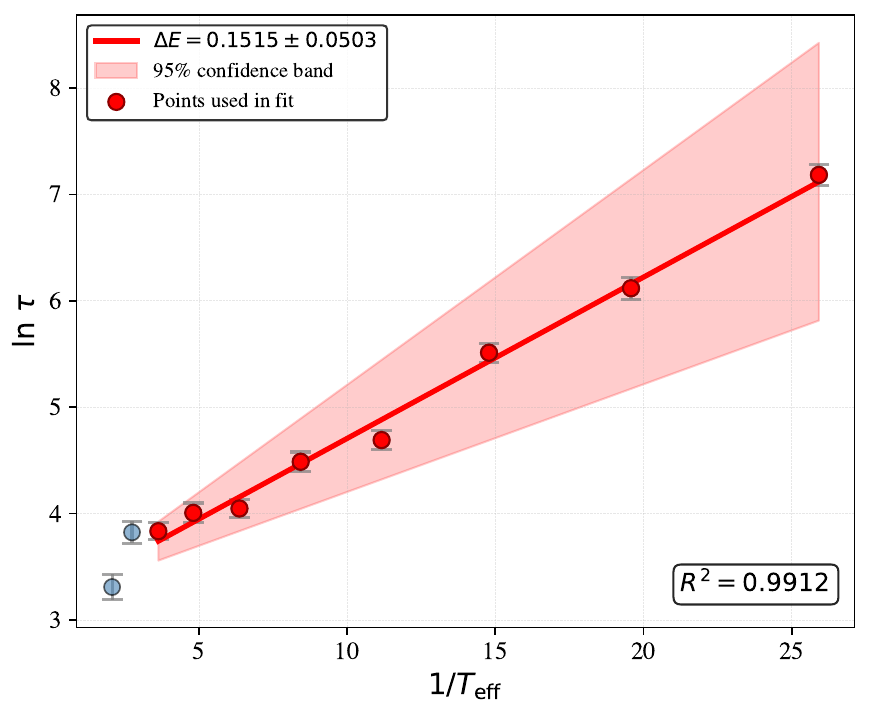}
   \par\vspace{0pt}\small\textbf{(b) Arrhenius Fit for Escape Times}
 \end{minipage}
 \caption{\textbf{(a)} Loss landscape section at $\beta=0.32$, $\eta=0.8$, $L=3$; dotted curves show $\beta=0.29$ and $\beta=0.35$. The barrier from the local minimum at $\lambda=0$ to the saddle gives $\Delta E_{\mathrm{min}}\approx0.003$. 
 \textbf{(b)} Arrhenius fit: $\ln\tau$ versus $1/T_{\mathrm{eff}}$. The linear fit ($R^2=0.991$) confirms thermally activated barrier crossing; the slope gives $\Delta E_{\mathrm{eff}}=0.15\pm0.05 \gg \Delta E_{\mathrm{min}}$, with the discrepancy explained by entropic and curvature corrections in $D=170$ dimensions (Appendix~\ref{app:effective_barrier}).}
 \label{fig:barrier_arrhenius}
\end{figure}

To test whether the escape is governed by the activated barrier crossing as predicted by Eq.~\eqref{eq:arrhenius}, we measured escape times from the rank-1 trapped state at $\beta=0.32$ while varying $\eta_{\mathrm{lr}}\in[5\times10^{-4}, 5\times10^{-3}]$ with batch size fixed at $B=64$. With $T_{\mathrm{eff}}\propto\eta_{\mathrm{lr}}/B$, Eq.~\eqref{eq:arrhenius} predicts $\ln\tau$ to be linear in $B/\eta_{\mathrm{lr}}$. Figure~\ref{fig:barrier_arrhenius}(b) shows this Arrhenius plot; the linear fit achieves $R^2=0.991$, confirming Eq.~\eqref{eq:arrhenius}. The slope yields the effective barrier:
\begin{equation}
  \Delta E_{\mathrm{eff}} = 0.15 \pm 0.05.
  \label{eq:barrier_eff}
\end{equation}
This is far above $\Delta E_{\mathrm{min}}\approx0.003$ shown in Fig.~\ref{fig:barrier_arrhenius}(a). The discrepancy is expected: in a $D=170$-dimensional parameter space the Kramers--Langer
formula adds entropic and geometric corrections from the $D-1$ transverse directions,
driving $\Delta E_{\mathrm{eff}} \gg \Delta E_{\mathrm{min}}$
(Appendix~\ref{app:effective_barrier}).

%-----------------------------------------------------------------------
\section{Discussion}
\label{sec:discussion}
%-----------------------------------------------------------------------

\subsection{A Mechanistic Account of Grokking}

Our results establish three claims. First, first-order L2 phase transitions produce coexisting metastable states whose barriers trap models in low-accuracy phases. Second, escape from these states is an activated process governed by Arrhenius kinetics with $T_{\mathrm{eff}}\propto\eta_{\mathrm{lr}}/B$. Third,
deliberate trapping reproduces the hallmark features of grokking, long delay, abruptness, sensitivity to initialisation, and the train/test dissociation under sparse sampling, in an analytically tractable minimal model. This framework makes concrete, falsifiable predictions:
\begin{enumerate}
  \item \emph{Staged grokking}: In tasks with $d$ learnable features, grokking should proceed in up to $d$ discrete stages, one per singular value of $\Sigma_{yx}$.
  \item \emph{Depth dependence}: Deeper networks ($L$ larger) produce higher barriers and thus longer grokking delays, since the critical strength $\beta_c^{(i)}$ decreases with $L$ while the equal-loss crossing $\beta^*_i$ remains finite~\cite{ladewig2025linear}.
  \item \emph{Hyperparameter control}: Escape time obeys $\ln\tau\propto B/\eta_{\mathrm{lr}}$, providing a direct lever to accelerate or suppress grokking.
\end{enumerate}
Predictions (2) and (3) are immediately testable by varying $L$, $\eta_{\mathrm{lr}}$, and $B$ in existing grokking benchmarks. Preliminary experiments and work on nonlinear networks ~\cite{ersoy2025hierarchical} with sigmoid and tanh activations reproduce qualitatively identical first-order phase transition behaviour, strongly suggesting that the same mechanism applies beyond the linear framework.

\subsection{Relation to Memorisation and Generalisation}
In the dense-data experiments of Fig.~\ref{fig:mse}(a) the training error converges monotonically, and the large train/test gap of canonical grokking~\cite{power2022grokking} does not appear, because with $d=2$ well-sampled modes the model has few directions along which train and test can diverge. Under sparse sub-sampling this gap does emerge: Fig.~\ref{fig:mse}(b) shows train error settling onto a plateau while test error remains substantially higher, until both fall abruptly as the model escapes the rank-1 phase. Because the network is linear and the task is stochastic, the model cannot memorise the training sample: train error plateaus at the best rank-1 linear fit and, after escape, settles near the best rank-2 fit, which is bounded below by the irreducible (Bayes) error of the Gaussian task. Train and test therefore approach one another and the noise floor but cannot coincide or fall to zero. The grokking signature here is thus the delayed, abrupt closing of a train/test gap down to the task's noise floor, reproduced \emph{without any memorising solution} — consistent with the view that memorisation is not a necessary ingredient of the phenomenon. When $d$ is large, the optimisation trajectory reduces training error by learning some singular directions while stagnating in others, giving rise to an apparent memorisation phase. In this view, memorisation and generalisation need not be primitive concepts but instead emerge as descriptions of partial versus complete progress through a cascade of rank transitions. Moreover, what is currently labeled ``grokking'' possibly encompasses several mechanistically distinct phenomena with a superficial resemblance; our framework offers a principled basis for distinguishing them.

%-----------------------------------------------------------------------
\section{Conclusion}
\label{sec:conclusion}
%-----------------------------------------------------------------------

We propose a candidate mechanism for grokking: noise-activated escape from metastable states created by first-order phase transitions in L2-regularised deep networks. Using deep linear networks, i.e. the minimal setting in which the loss landscape is exactly solvable~\cite{ladewig2025linear}, we demonstrate three results: \textbf{(i)}~deliberate trapping in metastable phases reproduces grokking-like delayed convergence; \textbf{(ii)}~escape times obey Arrhenius scaling $\ln\tau\propto 1/T_{\mathrm{eff}}$ with $T_{\mathrm{eff}}\propto \eta_{\mathrm{lr}}/B$; and \textbf{(iii)}~the extracted barrier $\Delta E_{\mathrm{eff}}=0.15\pm0.05$ exceeds the minimum-path barrier $\Delta E_{\mathrm{min}}\approx0.003$ by the factor predicted from entropic and curvature corrections in $D=170$ dimensions. The framework connects grokking to the established physics of noise-activated escape from metastable states~\cite{kramers1940,hanggi1990,mori2022sgd}, and offers a direct practical handle: since $\ln\tau\propto B/\eta_{\mathrm{lr}}$, grokking delays can in principle be accelerated or suppressed by hyperparameter choice alone.
\vspace{-0.8em}

%-----------------------------------------------------------------------
\section*{Acknowledgments}
We thank Bj\"orn Ladewig and members of the Complexity Science Group for stimulating discussions and inspiring contributions.

%-----------------------------------------------------------------------

%-----------------------------------------------------------------------

\newpage
\appendix

\section{Experimental Setup}
\label{app:experiments}

All experiments use a deep linear network of depth $L=3$ (two hidden layers) with hidden width $w=10$ and output dimension $p=2$, giving $D=170$ total parameters. No nonlinear activation is applied; the end-to-end map is a single matrix product. The optimiser is SGD throughout.

\textbf{Data generation.} Training and test data each consist of $N=512$ samples drawn from a zero-mean multivariate Gaussian over the joint input--output space. The data covariance has $d=2$ non-zero singular values $\eta_1=0.9$, $\eta_2=0.8$ with $\Sigma_{xx}=\mathbf{I}$. Samples are generated via Cholesky decomposition: given $\Sigma = LL^\top$, each sample is drawn as $z = L\epsilon$ with $\epsilon\sim\mathcal{N}(0,\mathbf{I})$; the first $p$ components are input and the remaining $p$ components are output. An independent test set of the same size is held fixed throughout.

\textbf{Initialisation and annealing.} All trapping experiments use \emph{checkpoint initialisation}: the model is first trained to convergence  $\beta_{\mathrm{init}} = 0$ to  get the full rank solution. The resulting weights are then used as the starting point for continued training. This is analogous to annealing. The random initialisation baseline uses fresh weights drawn from $\mathcal{N}(0, 1/\sqrt{w})$ instead. Convergence is assessed by monitoring test MSE; escape from a metastable phase is identified as the epoch at which test MSE drops below a threshold midway between the metastable plateau value and the global minimum value.

\textbf{Phase diagram} (Fig.~\ref{fig:phase_diagram}, Appendix~\ref{app:phase_diagram}): the model is first trained to convergence at $\beta=0$, then $\beta$ is increased quasi-statically in small increments, re-training to convergence at each step.

\textbf{Trapping experiments} (Fig.~\ref{fig:mse}): all three protocols use
$\beta=0.32$, $\eta_{\mathrm{lr}}=0.08$, $B=64$.
\begin{itemize}
  \item \emph{Random initialisation}: weights drawn from a standard normal distribution
        scaled by $1/\sqrt{w}$; training run for $2\times10^4$ epochs.
  \item \emph{Rank-1 trap}: model initialised from a checkpoint trained to convergence
        at $\beta=0.361>\beta_c^{(1)}=0.360$, placing the end-to-end weight matrix in the
        rank-1 phase; training then continued at $\beta=0.32$ for $2\times10^4$ epochs.
  \item \emph{Trivial-phase trap}: model initialised from a checkpoint trained to
        convergence at $\beta=0.43>\beta_c^{(2)}=0.42$, placing the network in the
        rank-0 phase; training continued at $\beta=0.32$ for $2\times10^4$ epochs.
\end{itemize}
Each experiment is repeated across 50 random seeds (seeds 42--91); escape times vary across seeds due to stochastic SGD noise, which is the activated escape mechanism under study.
\textbf{Arrhenius sweep} (Fig.~\ref{fig:barrier_arrhenius}(b)): starting from the rank-1 trap initialisation above, escape times are measured at $\beta=0.32$ with $B=64$ fixed while varying $\eta_{\mathrm{lr}} \in \{5\times10^{-4},\,1\times10^{-3},\,2\times10^{-3},\, 3\times10^{-3},\,5\times10^{-3}\}$. Each point is averaged over 50 seeds. The proportionality constant $C$ in $T_{\mathrm{eff}}=\eta_{\mathrm{lr}}/(BC)$ is estimated using the trace of the covariance of the gradients. The linearity ($R^2=0.991$) is a key result. With $N=512$ and $B=64$, each epoch consists of $\lfloor N/B \rfloor = 8$ SGD steps; this factor enters $\ln\tau_0$ but not the slope $\Delta E_{\mathrm{eff}}/T_{\mathrm{eff}}$, leaving the extracted barrier unchanged. 
\textbf{Sparse sub-sampling experiment} (Fig.~\ref{fig:mse}(b)): train and test are drawn from a single zero-mean Gaussian with weak correlation $0.8$ and strong correlation $0.9$ ($\Sigma_{xx}=\mathbf{I}$). The training set is a random $25$-sample subset of a $5\,000$-sample pool ($0.5\%$), so the weak feature is poorly determined from the training data while remaining at full strength on the (disjoint, $10\,000$-sample) test set. The model is initialised in the rank-1 trapped state and trained at $\beta=0.0025$, $\eta_{\mathrm{lr}}=0.1$, $B=64$. Escape is identified as the epoch at which the second singular value of the end-to-end map rises above a small threshold ($\sigma_2 > 10^{-3}$). Because the network is linear, the training error cannot reach zero: it plateaus at the best rank-1 linear fit and, after escape, settles near the best rank-2 fit, both bounded below by the irreducible (Bayes) error of the stochastic task.

%-----------------------------------------------------------------------
\section{Phase Diagram}
\label{app:phase_diagram}
\begin{figure}[h]
  \centering
  \includegraphics[width=0.8\columnwidth]{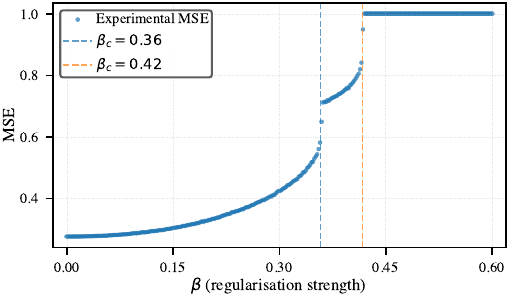}
  \par\vspace{0pt}\small\textbf{Experimental MSE vs Regularization Strength}
  \caption{Phase transitions in a deep linear network ($L=3$, $d=2$, $\eta_1=0.9$, $\eta_2=0.8$, $\Sigma_{xx}=\mathbf{I}$). Test MSE versus regularisation strength $\beta$ shows two sharp drops at $\beta_c^{(1)}=0.36$ and $\beta_c^{(2)}=0.42$; at each transition the effective rank increases by one. Quasi-static sweep; small offset from theory reflects output distribution bias.}
  \label{fig:phase_diagram}
\end{figure}

Figure~\ref{fig:phase_diagram} shows the test MSE as a function of $\beta$ for a deep linear network with $d=2$ singular values ($\eta_1=0.9$, $\eta_2=0.8$, $\Sigma_{xx}=\mathbf{I}$, $L=3$), obtained by training to convergence and then quasi-statically decreasing $\beta$ in steps of $0.01$. Two sharp drops mark the transitions at $\beta_c^{(1)}=0.36$ and $\beta_c^{(2)}=0.42$, in close agreement with Eq.~\eqref{eq:beta_c}. At each transition the effective rank of the end-to-end weight matrix increases by one, corresponding to the model learning an additional feature. The small offset from the theoretical values reflects the bias of the output distribution.

%-----------------------------------------------------------------------
\section{Bifurcation Diagram}
\label{app:bifurcation}
\begin{figure}[h]
  \centering
  \includegraphics[width=0.8\columnwidth]{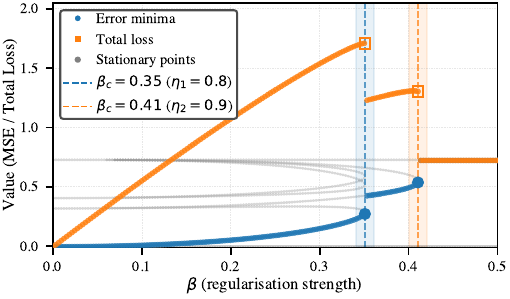}
  \par\vspace{0pt}\small\textbf{Numerical MSE and Total Loss vs Regularization Strength}
  \caption{Bifurcation diagram for modes ($\eta_1=0.9$, $\eta_2=0.8$, $L=3$). Solid lines: stable stationary points; dashed line: unstable saddle. The two non-trivial branches annihilate at $\beta_c$ in a saddle-node bifurcation, leaving only $\lambda=0$ for $\beta>\beta_c$. A model on the rank-1 branch (dotted continuation) remains metastable until stochastic noise drives it across the barrier into the rank-2 global minimum.}
  \label{fig:bifurcation}
\end{figure}

The bifurcation structure of the loss landscape is shown in Fig.~\ref{fig:bifurcation}. For a single mode with signal strength $\eta$, the stationarity condition Eq.~\eqref{eq:stationary} has three solutions below $\beta_c$: the trivial minimum at $\lambda=0$, an unstable saddle at $\lambda^{\mathrm{sad}}$, and the global minimum at $\lambda^{**}>\lambda^{\mathrm{sad}}$. As $\beta$ increases through $\beta_c$, the saddle and the global minimum merge and annihilate in a saddle-node bifurcation, leaving only $\lambda=0$. A model on the upper (rank-1) branch remains metastable for all $\beta<\beta_c$; a model on the lower (rank-0) branch at $\lambda=0$ is trapped there even when the rank-1 phase is energetically preferred (i.e., for $\beta<\beta^*$). The height of the barrier between the two, and thus the mean escape time, grows as $\beta$ increases toward $\beta_c$. Note that for $L=2$ no such bifurcation occurs: the stationarity condition is linear in $\lambda$ and has a unique positive solution for all $\beta>0$, yielding only second-order behaviour~\cite{ziyin2023zeroth}.

%-----------------------------------------------------------------------
\section{Critical Regularisation Strength}
\label{app:critical_beta}
The stationary condition for mode $i$ under the loss of Eq.~\eqref{eq:loss} is
\begin{equation}
  \frac{\partial\mathcal{L}}{\partial\lambda_i}
  = \lambda_i - \eta_i + \beta\,\lambda_i^{2/L-1} = 0.
  \label{eq:stationary_app}
\end{equation}

\subsection{Saddle-node bifurcation for DNNs}
Setting $k = 2/L$, the stationarity condition reads
\begin{equation}
  \lambda_i - \eta_i + \beta\,\lambda_i^{k-1} = 0.
  \label{eq:stat_k}
\end{equation}
For a range of $\beta$, this equation admits two distinct positive solutions: a larger root $\lambda^{**}$ (global minimum) and a smaller root $\lambda^{\mathrm{sad}}$ (unstable saddle). The two roots coalesce at $\beta_c^{(i)}$, requiring both Eq.~\eqref{eq:stat_k} and its derivative to vanish simultaneously:
\begin{equation}
  1 + \beta(k-1)\lambda_i^{k-2} = 0.
  \label{eq:second_deriv}
\end{equation}
Solving Eq.~\eqref{eq:second_deriv} gives the critical singular value $\lambda_c^{(i)} = \eta_i(1-k)/(2-k)$, and substituting back into Eq.~\eqref{eq:stat_k} yields Eq.~\eqref{eq:beta_c} of the main text.

\subsection{Number of transitions}
Because the modes decouple, each non-zero singular value $\eta_i$ of $\Sigma_{yx}$ has its own $\beta_c^{(i)}$. The ordering $\eta_1>\cdots>\eta_d>0$ implies $\beta_c^{(1)}>\cdots>\beta_c^{(d)}$, giving $d$ separate saddle-node bifurcations as $\beta$ is decreased from a large value.

%-----------------------------------------------------------------------
\section{Theoretical Barrier Heights}
\label{app:barriers}
For a single mode at $\beta < \beta_c^{(i)}$, the barrier for escape from the metastable lower rank phase is the loss difference between the saddle and the trivial minimum:
\begin{equation}
  \Delta E
  = \mathcal{L}(\lambda^{\mathrm{sad}}) - \mathcal{L}(0)
  = \tfrac{1}{2}(\lambda^{\mathrm{sad}} - \eta)^2
    + \frac{L\beta}{2}(\lambda^{\mathrm{sad}})^{2/L}
    - \tfrac{1}{2}\eta^2,
  \label{eq:barrier}
\end{equation}
where $\lambda^{\mathrm{sad}}$ is the smaller positive root of Eq.~\eqref{eq:stat_k}, obtained numerically. At $\beta = 0.32$, $\eta_2 = 0.8$, $L = 3$: $\lambda^{\mathrm{sad}} \approx 0.41$ and $\Delta E \approx 0.003$. This is the minimal barrier along the lowest-loss path out of the metastable phase. The large discrepancy with the experimentally extracted $\Delta E_{\mathrm{eff}} = 0.15 \pm 0.05$ is explained by the high-dimensional corrections derived in Appendix~\ref{app:effective_barrier}.

%-----------------------------------------------------------------------
\section{Effective Barrier Height in High-Dimensional Landscapes}
\label{app:effective_barrier}

\subsection{Kramers--Langer Theory}
In a parameter space of dimension $D$, the mean first-passage time from a metastable
minimum to a saddle point under Langevin dynamics with noise strength $T_{\mathrm{eff}}$
is given by the Kramers--Langer formula~\cite{hanggi1990}:
\begin{equation}
  \tau = \frac{2\pi}{|\omega_1^{\mathrm{sad}}|}
    \left(
      \frac{\prod_{j=1}^{D}|\omega_j^{\mathrm{sad}}|}
           {\prod_{j=1}^{D-1}\omega_j^{\mathrm{min}}}
    \right)^{\!\!1/2}
    \exp\!\left(\frac{\Delta E}{T_{\mathrm{eff}}}\right),
  \label{eq:kramers_langer_app}
\end{equation}
where $\omega_j^{\mathrm{min}} > 0$ are Hessian eigenvalues at the metastable minimum and $\omega_j^{\mathrm{sad}}$ those at the saddle, with exactly one negative eigenvalue $\omega_1^{\mathrm{sad}} < 0$.

\subsection{Effective barrier and entropic contributions}
Taking the logarithm of Eq.~\eqref{eq:kramers_langer_app} and defining
\begin{equation}
  \Delta E_{\mathrm{eff}}
  \equiv \Delta E
    - \frac{T_{\mathrm{eff}}}{2}
      \sum_{j=1}^{D-1}
        \ln\!\frac{\omega_j^{\mathrm{sad}}}{\omega_j^{\mathrm{min}}},
  \label{eq:eff_barrier_app}
\end{equation}
the Arrhenius form is recovered exactly: $\ln\tau = \mathrm{const} + \Delta E_{\mathrm{eff}}/T_{\mathrm{eff}}$. In our setting $D=170$. The metastable minimum is strongly confining in all directions, while the saddle has broader curvature in the $D-1$ transverse directions, so $\omega_j^{\mathrm{min}} \gg \omega_j^{\mathrm{sad}}$ for most $j$. The resulting entropic contribution drives $\Delta E_{\mathrm{eff}} \gg \Delta E_{\mathrm{min}}$, explaining the observed factor of $\sim 50$.

\subsection{Correction from multiplicative SGD noise}
Mori et al.~\cite{mori2022sgd} showed that for MSE loss, the multiplicative nature of SGD noise modifies the relevant barrier from the linear loss difference to a logarithmised quantity. For the approximately quadratic landscape near our metastable minimum, this correction does not alter the linear Arrhenius relationship, consistent with $R^2=0.991$.

\subsection{Escape time and choice of time unit}
Escape times $\tau$ are reported in epochs. Since $B$, $\eta_{\mathrm{lr}}$, and $N$ are fixed within each sweep, the conversion
between epochs and SGD steps is a constant (see Appendix~\ref{app:experiments}), leaving the extracted barrier height unchanged.

\section{Generalisation to Nonlinear Networks}
\label{app:nonlinear}
While our quantitative results (barrier heights, critical $\beta$ values) are specific to linear networks, the qualitative mechanism—metastable trapping due to saddle-node bifurcations, followed by noise-activated escape—applies generally. Prior work has established that deep nonlinear networks exhibit identical first-order L2-driven phase transitions~\cite{ziyin2023zeroth,ersoy2025curvature}, with hierarchical feature learning~\cite{ersoy2025hierarchical}. The main ingredient for proposing the hysteresis is the underlying first-order phase transition that exists beyond the linear setup. The linear case thus serves as a minimal model that preserves the essential bifurcation structure while remaining analytically tractable.
\end{document}